\documentclass[letterpaper, 10 pt, conference]{ieeeconf}  

\IEEEoverridecommandlockouts                              

\usepackage{epsfig} 
\usepackage{times} 
\usepackage{amsmath} 
\usepackage{amssymb}  

\usepackage{bm}
\usepackage{url}
\usepackage{nccmath}
\usepackage[utf8]{inputenc}

\usepackage[usenames] {color}

\newcommand{\optional}[1]{}

\title{\LARGE \bf
An Electromagnetism-Inspired Method for Estimating \\ In-Grasp Torque from Visuotactile Sensors
}

\author{Yuni Fuchioka$^{1*}$ and Masashi Hamaya$^{1}$
\thanks{$^{1}$OMRON SINIC X Corporation, Tokyo, Japan. $^{*}$Work done as an intern.}
}

\begin{document}

\maketitle
\thispagestyle{empty}
\pagestyle{empty}

\begin{abstract}

Tactile sensing has become a popular sensing modality for robot manipulators, due to the promise of providing robots with the ability to measure the rich contact information that gets transmitted through its sense of touch.
Among the diverse range of information accessible from tactile sensors, torques transmitted from the grasped object to the fingers through extrinsic environmental contact may be particularly important for tasks such as object insertion. However, tactile torque estimation has received relatively little attention when compared to other sensing modalities, such as force, texture, or slip identification.
In this work, we introduce the notion of the Tactile Dipole Moment, which we use to estimate tilt torques from gel-based visuotactile sensors. This method does not rely on deep learning, sensor-specific mechanical, or optical modeling, and instead takes inspiration from electromechanics to analyze the vector field produced from 2D marker displacements.
Despite the simplicity of our technique, we demonstrate its ability to provide accurate torque readings over two different tactile sensors and three object geometries, and highlight its practicality for the task of USB stick insertion with a compliant robot arm.
These results suggest that simple analytical calculations based on dipole moments can sufficiently extract physical quantities from visuotactile sensors.

\end{abstract}

\section{INTRODUCTION}

Tactile sensors have gained great interest in recent years as a promising sensing modality for robotic manipulation due to the variety of information that they can provide to robots, ranging from binary contact, pressure distribution, slip, vibration, texture, hardness, and among many others~\cite{yamaguchi-survey}.
Of the diverse range of existing tactile sensors, visuotactile sensors such as the Gelsight Mini \cite{gelsight-mini} or the DIGIT \cite{digit} have gained particular popularity due to their simple construction, high resolution, and ability to leverage techniques from computer vision, leading to wide commercial availability \cite{gelsight, deltact}. As a result, visuotactile sensors have demonstrated their ability to endow robot manipulators with a wide variety of tasks ranging from grasp stability prediction~\cite{si2022grasp}, touch-based object manipulation~\cite{tian2019manipulation}, or texture recognition~\cite{cao2020spatio}.

\begin{figure}[t]
    \centering
    \includegraphics[width=1.0\columnwidth]{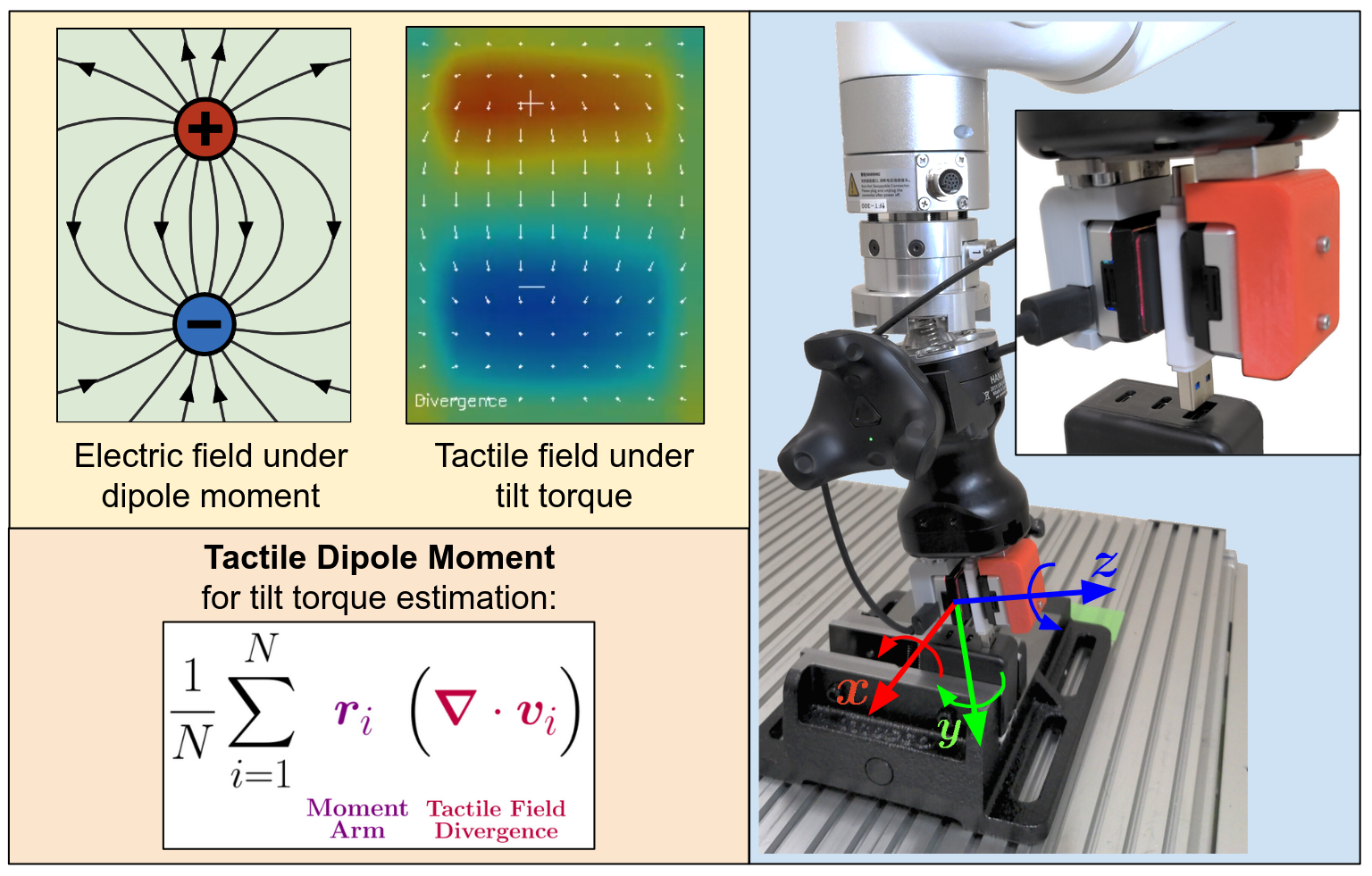}
    \caption{The core contributions of this work. Left, top: Similarities between the electric fields produced from a dipole charge distribution, and visuotactile marker motion fields produced from tilt torques. Left, bottom: The corresponding equation defining the tactile dipole moment used in this work to estimate tilt torques. Right: The contact-rich USB stick insertion alignment problem that we use to demonstrate our method.}
    \label{figure:fig-1}
\end{figure}

In this work, we are interested in the problem of using visuotactile sensors to estimate external forces and torques applied to grasped objects. In particular, tasks such as the contact-rich object insertion task illustrated in Fig. \ref{figure:fig-1} may require compliant interactions between the grasped object and the environment, as well as the ability to measure torques transmitted from the object to the gripper that result from this contact. However, the literature on estimating torques from visuotactile sensors is relatively limited. This is due to the limited ability of a typical visuotactile sensor consisting of a single gel and camera without additional hardware to provide information about gel displacement normal to the gel surface ($z$ axis as shown in Fig. \ref{figure:fig-1}) with the commonly used technique of the marker displacement method \cite{mdm-review}.

Zhang et al. \cite{tactile-nhhd} proposed to use a technique from flow analysis to relate patterns in the marker motion vector field, namely the diverging, unidirectional, and rotational components, to 3D forces and the 1D in-plane torques applied on the tactile gel. Inspired by this, we measure the remaining 2 dimensions of torque (axes $x$ and $y$ in Fig. \ref{figure:fig-1}), called the ``tilt" torque in \cite{gelsight}, using analysis techniques inspired by electrodynamics \cite{griffiths}. The core underlying idea comes from the observation that the marker motion displacement field that results from tilt torques closely resembles the electric fields produced from electric dipole moments. Therefore, we propose to use the same calculation used to characterize electric dipole moments to similarly estimate tactile tilt torques, and demonstrate through experiments that this simple calculation can enable successful measurement of tilt torques without relying on deep learning or sensor-specific analytical modeling of mechanical or optical properties.

The summary of our contributions is as follows.
\begin{itemize}
    \item We introduce the Tactile Dipole Moment as a method for estimating in-grasp tilt torques from visuotactile sensors, and demonstrate the improvement in estimation accuracy as compared to an existing analytical technique for tactile tilt torque estimation.
    \item Experiments of a USB stick insertion task with a real robot show that our estimation can provide useful feedback signals regarding environmental contact.
    \item Our method can be applied to other visuotactile sensing hardware and grasped object shapes.
    \item We open source the implementation of our algorithm and experimental apparatus design files\footnote{\url{https://github.com/omron-sinicx/tactile-dipole-monent}}.
\end{itemize}

\section{RELATED WORK}


\subsection{Marker Displacement Method}
The vector field that results from tracking displacements of markers or dense patterns on the gel surface through the Marker Displacement Method (MDM) \cite{mdm-review}, has become a popular data representation for visuotactile sensor outputs.
Using the terminology of \cite{mdm-review}, this method can be applied in 2D to consider only the translational motion of the markers within the camera \cite{gelsight-mini, digit, gelsight, deltact,  gelslim3.0, fingervision,  sferrazza2019},
or in 3D to obtain full information about the gel surface deformation in all spacial dimensions \cite{gelstereo, tac3d}.

Of the variety of visuotactile sensors that have been developed, those that rely primarily on 2D MDM have achieved commercial availability and wide adoption due to its simplicity and resulting ease of fabrication.
This is in comparison to sensors containing multiple cameras \cite{gelstereo}, special gel structures like pins \cite{tactip}, double layered embedded markers \cite{gelforce}, or mechanical structures in addition to the gel such as springs \cite{f-touch, visiflex}. 
Therefore, we wish to further develop algorithms to infer 3D data from widely available 2D sensors, despite its fundamental hardware limitations \cite{mdm-review}.


\subsection{Estimating Torques from Marker Displacements}
From the distributed forces encoded in the marker displacement vector field, the force and torque wrench at the contact point can be estimated \cite{gelsight}.
Within the full 6-dimensional force and torque wrench vector, the 2D shear force and 1D in-plane torque can be readily estimated from 2D marker motion displacement \cite{tactile-nhhd}.
More difficult to obtain are the remaining 1D normal force and 2D tilt torques, which require reasoning about distributed forces acting along the gel surface normal \cite{mdm-review}. 

To this end, \cite{fingervision} used the vector norm to estimate point-wise normal force, and combined this with point-wise shear force estimation via 2D marker displacement to calculate aggregate forces and torques. However, the vector norm cannot distinguish between marker motions exhibiting ``spreading" behaviors characteristic of normal forces, with that of shear displacements. This limitation is addressed in \cite{tactile-nhhd}, where diverging, unidirectional, and rotational patterns are explicitly decomposed through the natural Helmholtz-Hodge Decomposition \cite{nhhd}. This method provided estimation of 3D force and 1D torque along the gel surface normal, without 2D tilt torques. Additional results estimating wrenches involving normal forces include \cite{gelsight, densetact2}, which performed deep learning with large network architectures and large amounts of data, or \cite{extrinsic-contact-sensing}, which solved a closely related problem of 3D marker motion tracking through the inverse Finite Element Method (FEM) informing the depth estimation \cite{gelslim3.0, gelslim-inverse-fem}. In comparison to these approaches, our method estimates tilt torques using vector calculus approaches that do not involve deep learning or modeling of sensor properties.

\subsection{Tactile Sensing for Tool Manipulation}
As compared to a force/torque (FT) sensor mounted on a rigid manipulator robot, the small displacements of objects within the compliant gel fingertip of a visuotactile sensor provide improved measurements of extrinsic contact information between the grasped object and the environment that it interacts with \cite{extrinsic-contact-sensing, kim2022active, tactile-rl}.
This property has demonstrated its utility for tasks involving tool use \cite{fingervision, shirai2023tactile}, insertion \cite{kim2022active, tactile-rl, fu2023safe, zhao2023skill}, and for skills involving in-hand object manipulation \cite{digit, tian2019manipulation, wang2020swingbot}. Due to the relative simplicity of estimating in-plane torques, as discussed in the previous section, these works primarily relied on measuring rotation and torque about this axis. However, the ability to measure torques in the remaining two tilt axes may extend the capabilities of these results, and enable tasks such as the USB stick insertion illustrated in Fig. \ref{figure:fig-1}.

\section{METHOD}

\subsection{Problem Setting: In-Grasp Torque Estimation}
We consider the setting where an object is grasped by a parallel gripper with visuotactile sensors mounted on each finger, and this grasped object interacts with external forces and torques that must be sensed from the visuotactile sensors. We assume that the gripping force is sufficient so that the grasped object does not slip within grasp. The USB stick insertion problem shown in Fig. \ref{figure:fig-1} is an example of this situation.
Here, the torques transmitted from the object to the gripper are of particular importance out of the 6D force and torque wrenches due to the geometry of the moment arm induced by the grasped object.

\subsection{Tilt Torque Estimation via Tactile Dipole Moment} \label{section:method-dipole}
In \cite{tactile-nhhd}, it was introduced that different loading conditions on the tactile gel surface relate to the diverging, unidirectional, and rotational components of the marker displacement fields. Inspired by this insight, we make the additional observation that the marker displacement field pattern that results from tilt torques produces vector field patterns similar to electric fields induced by an electric dipole, which is defined as a charge distribution characterized by equal and opposite charge distributions separated by some (possibly infinitesimal) distance, as illustrated in Fig. \ref{figure:fig-1}.

This is because, in electrodynamics, the diverging component of the electric field relates to the charge distribution that induces it, through Gauss's Law \useshortskip
\begin{align}
    \bm{\nabla} \cdot \bm{E} = \frac{1}{\epsilon_0} \rho. \label{equation:gauss}
\end{align}
Here, $\bm{E}$ is the \textit{electric field}, which we draw analogies to the \textit{tactile marker displacement field}, $\rho$ is a position-dependent \textit{electric charge distribution}, which we relate to the \textit{normal force distribution} on the tactile gel, and $\epsilon_0$ is a constant known as the permittivity of free space \cite{griffiths}.
Analogously to the electric dipole moment, in the mechanical system of the tactile gel deformation under tilt torques, there are normal forces applied into and out of the gel surface in equal and opposite directions with the opposing regions separated by some distance. As per \cite{tactile-nhhd}, normal forces are related to the marker motion field divergence similarly to electric charge distributions relating to electric field divergence.

Therefore, we characterize the tilt torques applied on the gel surface using the same calculation used to characterize electric dipoles--the dipole moment $\bm{p}$, defined as \useshortskip 
\begin{align}
    \bm p = \iiint \bm{r} \rho \, dx \,dy\,dz,
\end{align}
where $\bm{r}$ is the distance vector from some predefined point in space to the infinitesimal volume element for integration \cite{griffiths}. We can additionally substitute Gauss's Law (\ref{equation:gauss}) to obtain the dipole moment in terms of the electric field \useshortskip
\begin{align}
    \bm p = \epsilon_0 \iiint \bm{r} \Big( \bm{\nabla} \cdot \bm{E} \Big) \, dx \,dy\,dz.
\end{align}
Transferring this calculation to the domain of tactile tilt torque estimation, we define the \textbf{tactile dipole moment} as \useshortskip
\begin{align}
    \bm p_{tilt} &= \frac{1}{N}\sum_{i=1}^{N} \bm{r}_i \Big( \bm{\nabla} \cdot \bm{v}_i \Big) \\
    &= \frac{1}{N} \sum_{i=1}^{N}
    \begin{bmatrix}
        (r_i)_x \\ (r_i)_y
    \end{bmatrix}
    \Bigg( \frac{d(v_i)_x}{dx} + \frac{d(v_i)_y}{dy} \Bigg),
    \label{equation:p-tilt}
\end{align}
where we replaced the volume integral with a summation over vectors $\bm{v}_i$ representing a discretized vector field indexed by $i \in [1, \ldots , N]$, and $\bm{r}_i$ is the moment arm vector.
Since the dipole moment is dependent on the origin from which to define $\bm {r}_i$ in the general case where the net charge is nonzero \cite{griffiths}, we take the integral from the midpoint between the centroids of the positive and negative divergence regions to approximate the point about which the planar gel surface rotates in reaction to tilt torques. Specifically, we first define the positive divergence mask function $\rho_i^+$ as having value $\bm{\nabla} \cdot \bm{v}_i$ if it is positive, and 0 otherwise.
Using this, the positive divergence $x$ axis centroid \useshortskip
\begin{align}
    C^+_x = \frac{\sum_i^N \big( \rho_i^+ x_i\big)}{\sum_i^N \rho_i^+}
\end{align}
can be calculated, where $x_i$ is the $x$-coordinate for marker $i$.
We can similarly calculate $C^+_y$, $C^-_x$, and $C^-_y$, using marker $y$ coordinates $y_i$, and with $\bm{\nabla} \cdot \bm{v}_i < 0$ as the condition for the negative divergence mask function.
From these, we define the centroid midpoint \useshortskip
\begin{align}
    \bm m = \frac{1}{2} \Big[ (C^+_x + C^-_x), (C^+_y + C^-_y) \Big]^T,
\end{align}
which is used to define \useshortskip
\begin{align}
    \bm r_i = \Big[ (x_i - m_x), (y_i - m_y ) \Big]^T.
\end{align}

Finally, since the dipole moment points from negative to positive charges, the torque $\bm{\tau}_{tilt}$ points in a direction perpendicular to this dipole moment and is scaled by constant calibration factors $c_x, c_y$, \useshortskip
\begin{align}
     \bm{\tau}_{tilt} = \Big[c_x (p_{tilt})_y , -c_y (p_{tilt})_x \Big]^T.\label{equation:tau-tilt}
\end{align}



\begin{figure}[t]
    \centering
    \includegraphics[width=1.0\columnwidth]{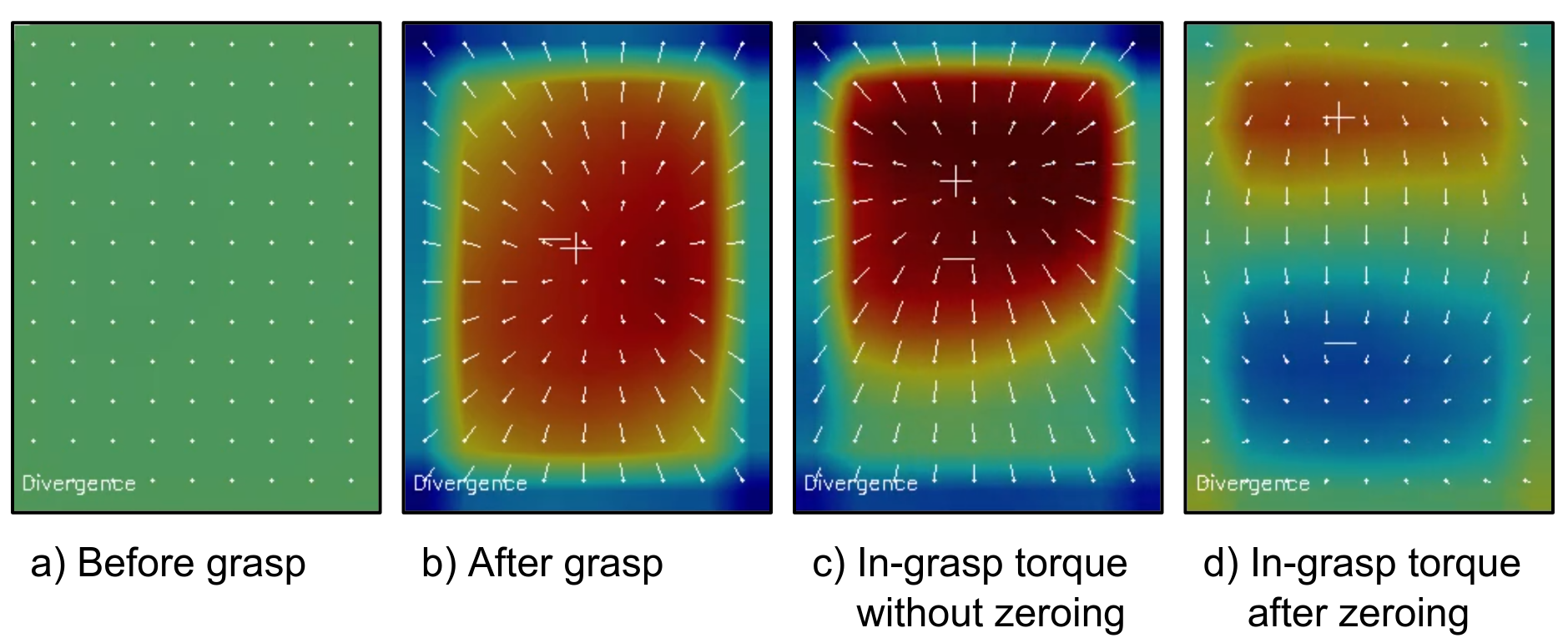}
    \caption{The diverging components of the tactile field \textit{before} and \textit{after} grasping an object respectively (a,b), and the field under the application of tilt torque \textit{with} and \textit{without} zeroing the tactile field after grasping, respectively (c,d). Note that the dipole field pattern only appears after zeroing the field after grasping. The diverging vector field component is obtained through \cite{nhhd}.}
    \label{figure:torque_zero}
\end{figure}

We remark that the dipole pattern only appears in the marker displacement fields when the vector field is zeroed after grasp, as illustrated in Fig. \ref{figure:torque_zero}.
In other words, the marker displacement measurements are taken by subtracting the post-grasp marker locations from the current marker locations in the image plane.
Without doing this, the displacement field is dominated by the component caused by the normal force, which remains relatively constant during grasp and makes detection of the tilt torques difficult due to its relatively small effect by comparison.

\section{EXPERIMENTS}

\subsection{Overview}
We evaluate our method through a series of experiments designed to characterize its capabilities and potential applications. In particular, we aim to 1) test the accuracy of the estimation within ideal conditions, 2) evaluate whether these estimates can be used for practical robot manipulation scenarios, and 3) analyze the generalization capability of our method across sensors and grasped object geometry. Note that for all experiments, we used the optical flow algorithm to generate vector fields rather than marker recognition and tracking, which is commonly used for sensors without dense optical patterns \cite{mdm-review}. This was a practical design decision to ease algorithm transfer across different sensor hardware.

%
\subsection{Main Result: Calibration and Evaluation} \label{section:main-result}
\subsubsection{Experimental Setup}
\begin{figure}[t]
    \centering
    \includegraphics[width=1.0\columnwidth]{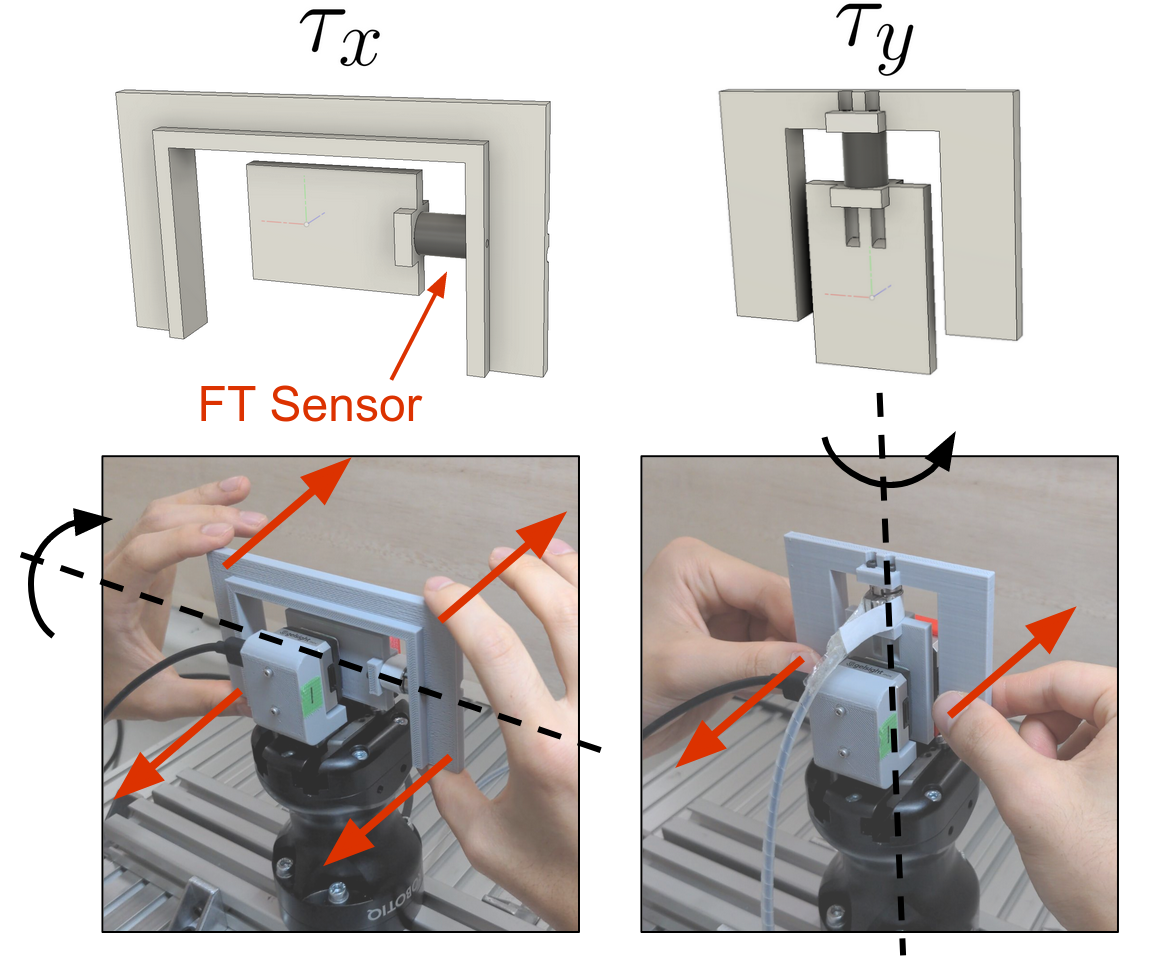}
    \caption{The experimental setup for our main calibration and evaluation experiments, showing the 3D printed jig to enable direct measurement of the torques applied to a grasped object (top), and the way in which a human manually applies the torque (bottom).}
    \label{figure:ft-plate}
\end{figure}

Using the method outlined in Section \ref{section:method-dipole}, we compare tilt torques estimated by our approach with that measured from a small and high resolution FT sensor (Nippon Liniax Corp., TFS12A-25), which we use as ground truth wrench measurements. The setup is shown in Fig. \ref{figure:ft-plate}. Unlike the commonly used setup involving a single visuotactile sensor with an FT sensor mounted underneath \cite{gelsight, tactile-nhhd, f-touch}, we are interested in measuring torques applied to an object while it is grasped by a robotic gripper. Therefore, the FT sensor is integrated into a plate grasped by the Robotiq Hand-E gripper, rather than being part of the gripper assembly. In the case of quasi-static loading, we expect the moment between the inner plate to the tactile gel to be equal to the moment applied from the outer plate to the inner plate through the FT sensor. Although only one Gelsight Mini visuotactile sensor \cite{gelsight-mini} is used to collect data, another unpowered sensor is attached to the other finger of the gripper to provide an elastic gel surface to enable the grasped object to displace in reaction to applied wrenches. We found that this angular compliance was key in reliably producing dipole field patterns, as compared to a setup involving a rigid finger on the other side of the tactile sensor used for measurement.

The gripper holds the plate with a fixed finger width, so that a human can manually apply tilt torques to the plate after zeroing the tactile field. During a single data collection trial, torques are applied along a single axis in alternating directions quasi-statically, with a typical torque profile applied during a data collection trial shown in Fig. \ref{figure:ft-time}. The single-axial nature of the applied wrenches is to avoid issues relating to incipient slip during this initial experiment, which is discussed in further detail in Section \ref{section:incipient-slip}.

This procedure is performed for several ranges of maximum torque values, producing a dataset consisting of 12,497 and 12,797 tactile sensor readings respectively for the $\tau_x$ and $\tau_y$ experiments, where the asynchronous and higher frequency data of the FT sensor, sampled at roughly 62.5Hz, was interpolated to match the timings of the tactile sensor outputs at roughly 19Hz to provide ground truth values for comparison.
As a baseline method that also estimates tilt torques from 2D visuotactile data without deep learning, the method described in \cite{fingervision} was employed. Rewriting with the notational convention used here, the point-wise force estimate in \cite{fingervision} is given by \useshortskip
\begin{align}
    \bm{f}_i = [c_x (v_i)_x, c_y (v_i)_y, c_z \sqrt{(v_i)_x^2 + (v_i)_y^2}]\label{equation:fingervision-force}
\end{align}
for $\bm{v}_i, c_x, c_y, c_z$ defined similarly as (\ref{equation:p-tilt}) and (\ref{equation:tau-tilt}).
Then, the torque estimate is given by taking the cross product between (\ref{equation:fingervision-force}) and the distance from the center of the tactile image $\bm{l}_i$, \useshortskip
\begin{align}
    \bm{\tau}_{baseline} = \frac{1}{N}\sum_{i=1}^{N} \big( \bm{l}_i \times \bm{f}_i \big).
\end{align}
The major differences between this calculation and ours are a) the moment arm is taken from the middle of the image rather than being adaptive to the force distribution, and b) the normal force $f_z$ is estimated through the vector norm rather than the divergence, which cannot differentiate between diverging, rotating, or translational field patterns.


\subsubsection{Results}

\begin{figure*}[thpb]
    \centering
    \includegraphics[width=1.0\textwidth]{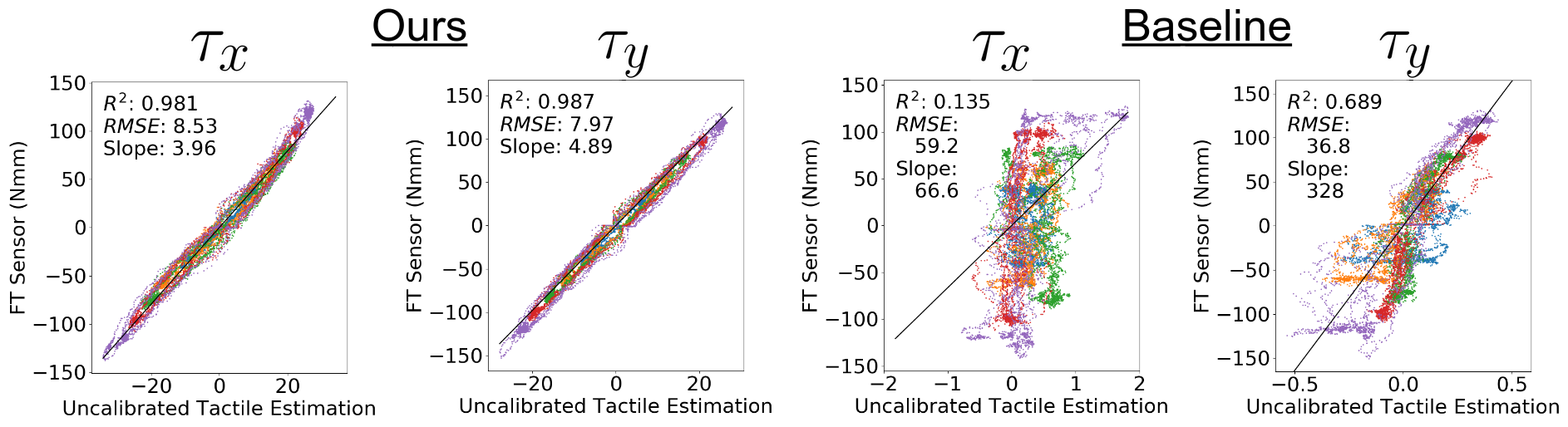}
    \caption{The main result of this work, showing the linear relationship between our method to estimate tilt torques, and the ground truth provided by the FT sensor (Left). This is compared against the method proposed in \cite{fingervision}, which also estimates tilt torques without the use of deep learning (right). The colors correspond to different experiments, each involving different maximum applied torques. Units for RMSE and Slope are in Nmm.}
    \label{figure:main-result}
\end{figure*}

\begin{figure}[t]
    \centering
    \includegraphics[width=1.0\columnwidth]{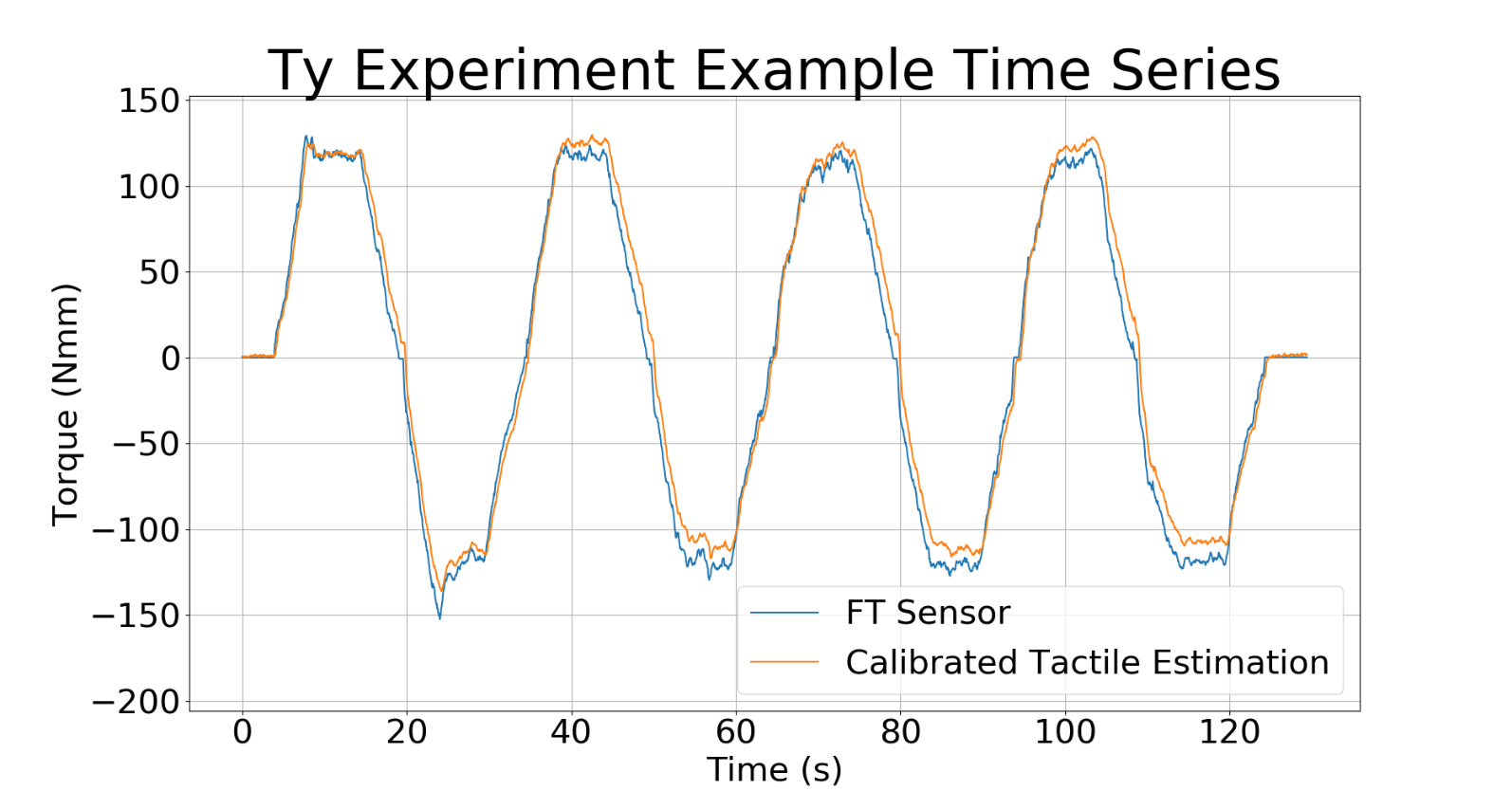}
    \caption{A typical time series plot of the torque applied during an experiment, both as obtained from the FT sensor as well as the estimation obtained from our method. This shows the quasi-static nature of the loads that we applied for evaluation. The scaling factor for the tactile estimation curve is obtained through the experiment illustrated in Fig. \ref{figure:main-result}.}
    \label{figure:ft-time}
\end{figure}



Fig. \ref{figure:main-result} shows the linearity between the FT sensor ground truth versus our estimation method, compared to the method given in \cite{fingervision}. Our method demonstrates higher estimation accuracy from the improved distributed normal force measurement via vector divergence over the vector norm. Note that the calibration scaling factor between the tactile estimation and FT sensor ground truth is roughly equal between the $x$ and $y$ axes, as is expected from the homogeneity of the gel material. We additionally observed that this linearity was maintained whenever torques along $x$ and $y$ axes were coupled, provided that the FT sensor was placed in line with the axes of rotation, to avoid creating moment arms causing human-applied torques to be read as forces in the FT sensor. Fig. \ref{figure:ft-time} shows the comparison between estimated and ground truth tilt torque values plotted with respect to time for a single data collection experiment, showing the practical application of the calibration experiment to produce accurate torque readings in engineering units from the tactile sensor.

\subsection{Application for USB Stick Insertion}\label{section:usb}
%
\begin{figure*}[thpb]
    \centering
    \includegraphics[width=1.0\textwidth]{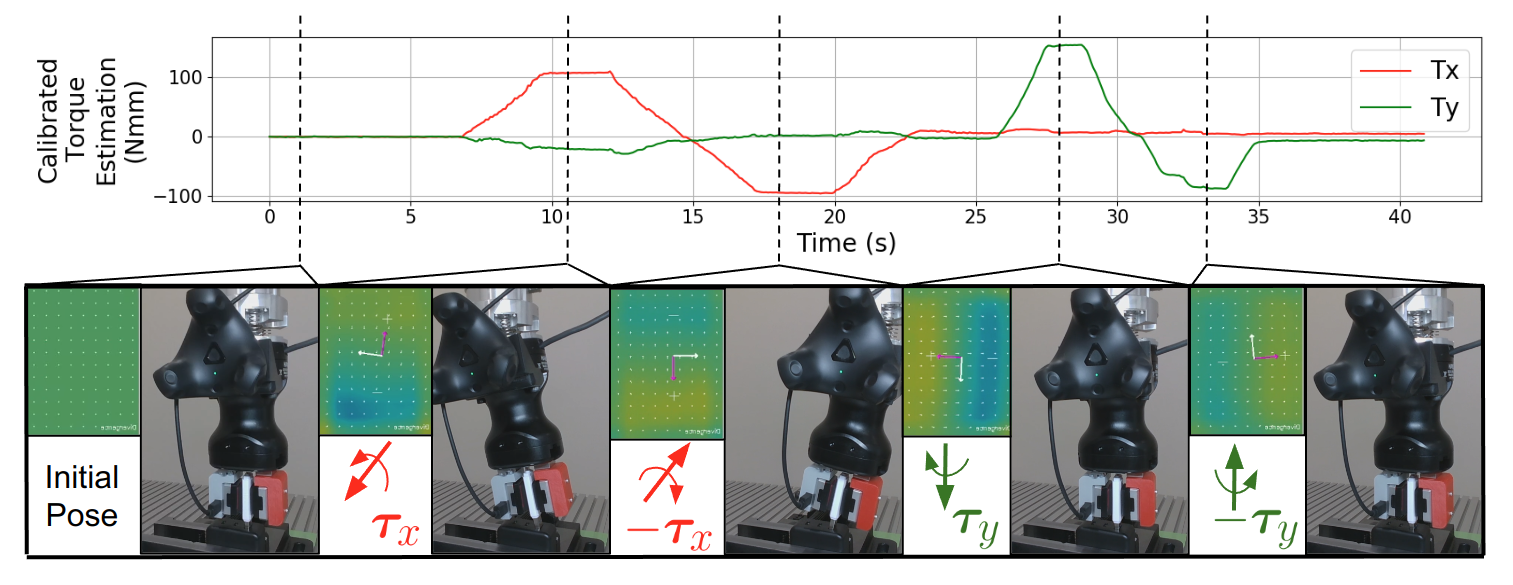}
    \caption{An application of our torque estimation method to the problem of measuring alignment of a USB stick for insertion, for a human-controlled robot manipulator equipped with a compliant wrist to enable contact-rich insertion \cite{soft-wrist}. The figure shows the time-series plot of the estimated torques (top), as well as representative video snapshots showing the state of the USB stick that caused the torque (bottom).}
    \label{figure:usb-alignment}
\end{figure*}
\subsubsection{Experimental Setup}
In order to demonstrate the utility of measuring tilt torques with our method, we consider an application of USB stick insertion alignment. The same parallel gripper with tactile sensors used in Section \ref{section:main-result} is now mounted on the UFACTORY xArm6 through the spring-loaded compliant wrist proposed in \cite{soft-wrist} to enable safe contact interactions. We track the pose of the compliant wrist with the VIVE Tracker motion capture device for visualization purposes.
Here, we consider the setting of a human teleoperating the robot through a gamepad controller rather than an automatic feedback controller, to manually command motions to focus the evaluation on our torque estimation method as well as to simplify the problem.

\subsubsection{Alignment measurement}
In the first experiment of alignment measurement, the robot starts from a pose where the USB stick is partially inserted into and aligned with the hole. Then the robot arm is commanded to translate about the $z$ axis and rotate about the $y$ axis according to the coordinate definition given in Fig. \ref{figure:fig-1}. Due to the compliance in the wrist, these misalignments result in tilt torques in $x$ and $y$ directions respectively, transmitted from the USB stick to the tactile fingers and estimated with our method. The results for both USB-A and USB-C connectors are shown in the supplemental video, as well as the snapshots given in Fig. \ref{figure:usb-alignment}. 


\subsubsection{Insertion}
In the second experiment of insertion, the robot starts from a state where the hole has some pose uncertainty relative to the initial USB stick pose. The tactile torque estimates aid in insertion through a) indicating when the hole is ``found" while dragging the USB stick along the surface, and b) informing when the USB stick is aligned with the hole when all torque estimates are small, indicating that a vertical motion will result in successful insertion. Since our method provides estimates for $x$ and $y$ axis tilt torques, the additional $z$ axis in-plane torque estimates are provided through the method proposed in \cite{tactile-nhhd}. The results for USB-A and USB-C insertion are provided in the supplemental video.



\subsection{DIGIT Sensor} \label{section:digit}
\begin{figure}[t]
    \centering
    \includegraphics[width=1.0\columnwidth]{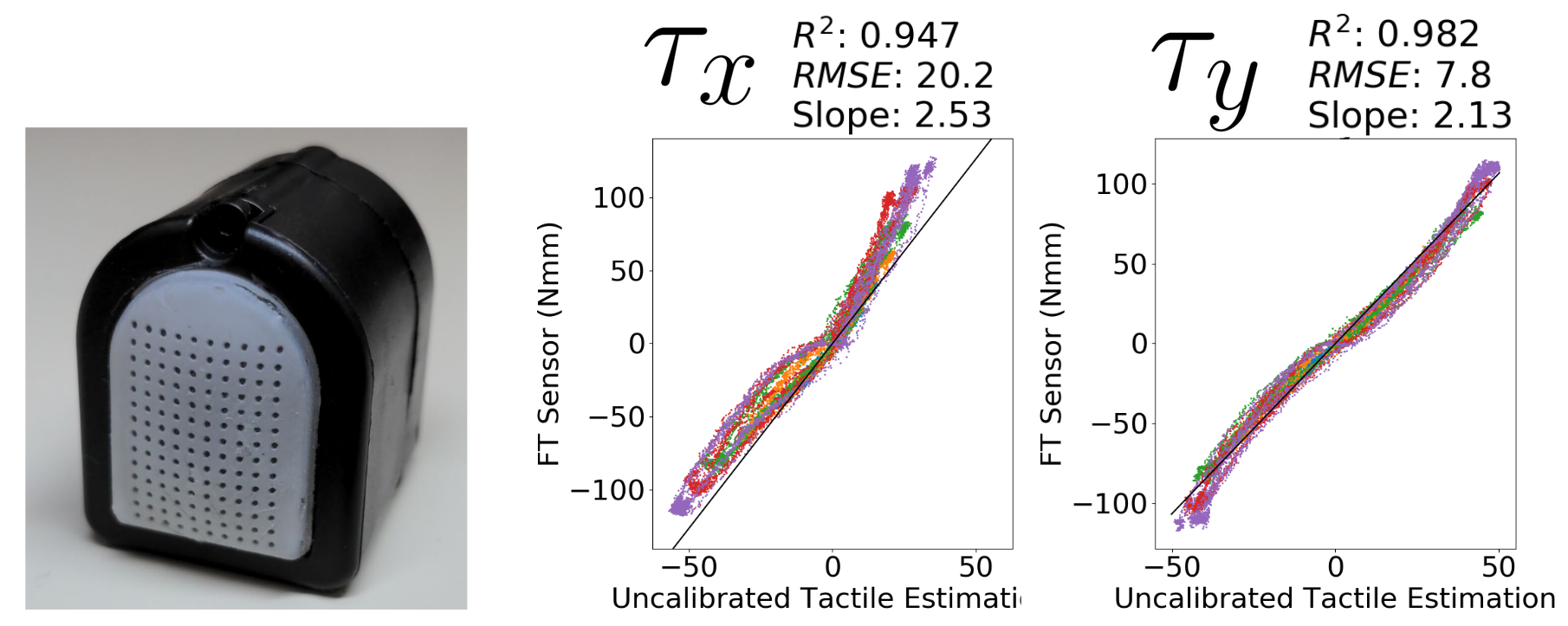}
    \caption{Experimental setup and results for the DIGIT sensor \cite{digit} transfer experiment. Left: The DIGIT sensor that we laser-etched to have markers, and Right: the results of comparing linearity with the FT sensor.}
    \label{figure:digit}
\end{figure}

Since our method is neither reliant on deep learning \cite{gelsight, densetact2} nor sensor-specific modeling of mechanical or optical properties \cite{gelforce, gelslim-inverse-fem, ground-truth-fem}, we expect our method to be directly applicable to any visuotactile sensor providing 2D marker displacement vector fields, with the only variation resulting from the calibration scaling constant from fitting a linear model to the ground truth FT sensor data. We test this by applying our method to the DIGIT sensor \cite{digit}, which we modified to include markers through laser etching as shown in Fig. \ref{figure:digit}, left. The results are provided in Fig. \ref{figure:digit}, right. We observe that the gel on the DIGIT sensor is stiffer and has a rounder surface profile compared to the Gelsight Mini, which means that tilt torques primarily cause a change in the contact surface to create non-uniform marker displacement motions, rather than the uniform planar motions as with the Gelsight Mini. Despite these hardware differences, these result demonstrates that the method can generalize across sensors, albeit with slightly reduced accuracy.

\subsection{Generalization to Object Shape}
\begin{figure}[t]
    \centering
    \includegraphics[width=1.0\columnwidth]{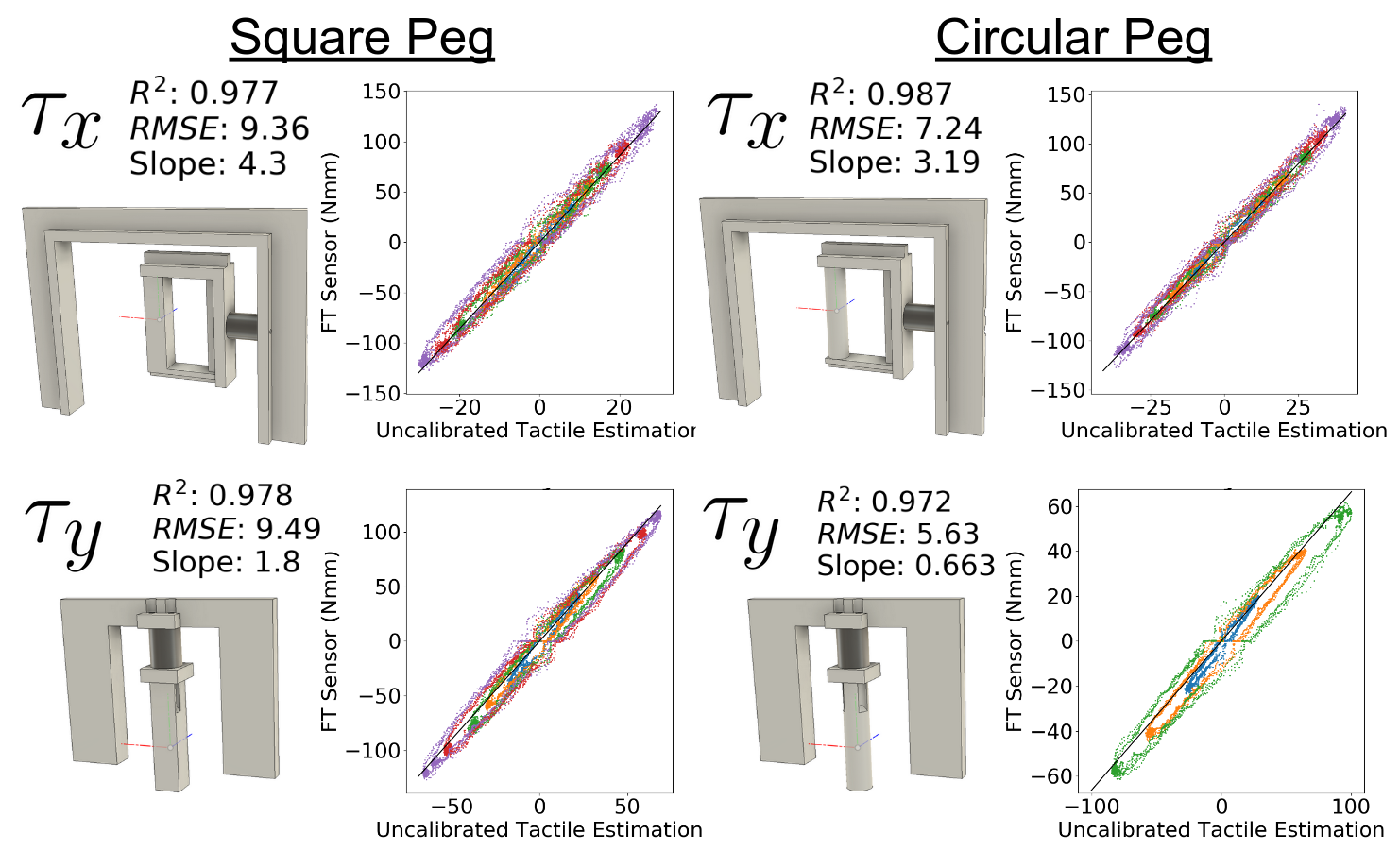}
    \caption{The 3D printed jigs used to test our method on peg-shaped objects (left), and experimental results (right). Only small torque values were evaluated for torques along the longitudinal axis of the circular peg ($\tau_y$), as significant slip occurred beyond the values shown here.}
    \label{figure:peg}
\end{figure}

The results shown in the previous sections were produced from the tactile fingers grasping objects with uniform flat regions.
To evaluate the generalization capability of our approach to object shapes, Fig. \ref{figure:peg} shows results for performing the same experiments using square and round pegs (10mm cross sectional width/diameter) rather than flat plates. We chose peg-shaped objects since the torque estimation problem is most relevant for grasping long objects that induce a moment arm when interacting with the environment. Due to the non-uniformity of the contact region, the calibration scaling factors are no longer consistent between the $x$ and $y$ axes, signifying that this calibration needs to be performed separately for every type of grasped object if a numerical torque estimate in engineering units is desired. Nonetheless, the linearity of data is conserved, showing the ability of our technique to generalize across object shapes.

\section{DISCUSSION AND LIMITATIONS}
The results presented in this work demonstrate how our method can be used to provide accurate tilt torque readings from simple analytical calculations. In this section, we discuss the implications and limitations of these results in relation to the relevant existing literature.

\subsection{Improved Analytical Estimation of Tilt Torques}
When compared to existing approaches for tilt torque estimation via analytical calculations from 2D marker displacement fields \cite{fingervision}, the results in Section \ref{section:main-result} demonstrate the improved measurement accuracy enabled through our approach. Although both methods are based on the principle of integrating torque contributions from normal forces distributed on the gel, the use of vector field divergence for normal force estimation, and the moment arm definition adapting to the force distribution location both serve to improve the estimation.
\subsection{Incipient Slip}\label{section:incipient-slip}
We identify incipient slip as a fundamental challenge for the in-grasp wrench estimation problem with visuotactile sensors, which is likely not limited to tilt torque estimation using our method.
In particular, we observe that large and quickly varying loads on grasped objects cause non-uniform gel deformations, resulting in erroneous estimates. After incipient slip occurs, the marker displacement field needs to be re-zeroed as described in Section \ref{section:method-dipole}.
We found that changing the grip strength also changes the calibration scaling constant between the tilt torque and tactile readings, making the necessary adaptive grip strategy, such as slip detection \cite{yamaguchi-survey, incipient-slip}, more complicated than simply increasing grip strength for our problem setting.
\subsection{Deep Learning}
The results presented in this paper rely only on analytical calculations without the use of deep learning.
This may be advantageous, as neural network training may require a large amount of data to achieve satisfactory performance, whereas simpler tactile representations can provide better generalization performance in the low data regime \cite{tactile-nhhd, tactile-rl}. On the other hand, given a sufficiently large training dataset, deep learning can be expected to outperform simple models.
As a third possibility, \cite{deltact, tactile-nhhd} demonstrated how effective analytical decompositions can reduce the complexity of data-driven models, which we also expect to hold for our method.
\section{CONCLUSION AND FUTURE WORK}
In this paper, we introduce the Tactile Dipole Moment as a technique to estimate tilt torque from visuotactile sensors. We test the efficacy of this simple method across two different sensor models and three grasped object geometries, and apply it to the problem of contact-rich USB stick insertion. Our results demonstrate the effectiveness of vector calculus techniques for analyzing visuotactile data. Future work could involve the solution of the tactile object insertion problem through automatic feedback control using our torque estimation method as sensory inputs.

\addtolength{\textheight}{-1cm}   





\section*{ACKNOWLEDGMENTS}
We wish to thank the engineers for their support in developing the necessary robot infrastructure, and for the members of the ETH Zürich Robotic Systems Lab for their valuable discussions and feedback.
This study is supported by JST ACT-X, Grant Number JPMJAX22AC.


\bibliographystyle{IEEEtran}
\bibliography{bibliography}

\end{document}